# Finding Stakeholder-Material Information from 10-K Reports using Fine-Tuned BERT and LSTM Models

Victor Chen (founder@gopeaks.org)

*Draft August 14, 2023*

## Abstract

All public companies are required by federal securities law to disclose their business and financial activities in their annual 10-K reports. Each report typically spans hundreds of pages, making it difficult for human readers to efficiently identify and extract the material information. To solve the problem, I have fine-tuned BERT models and RNN models with LSTM layers to identify stakeholder-material information, defined as statements that carry information about a company's influence on its stakeholders, including customers, employees, investors, and the community and environment. The existing practice is to use keyword search to identify such information, which is my baseline model. Using business expert-labeled training data of nearly 6,000 sentences from 62 10-K reports published in 2022, the best model has achieved an accuracy of 0.904 and an F1 score of 0.899 in test data, significantly above the baseline model's 0.781 and 0.749 respectively. Furthermore, the same work was replicated on more granular taxonomies, based onwhich four distinct groups of stakeholders (i.e., customers, investors, employees, and the community/natural environment) are tested separately. Similarly, fined-tuned BERT models outperformed LSTM and the baseline. The implications for industry application and ideas for future extensions are discussed.

**Acknowledgements:** The author owes thanks to Jennifer Zhu (AWS) for comments. This research was conducted independently from the author's affiliation. The codes and data can be found at https://github.com/GoPeaks-AI/stakeholder_material_reader.git

## Introduction

As federal securities law requires, an annual report on Form 10-K provides a comprehensive overview of the company's business and financial condition and includes audited financial statements. The annual 10-K report intends to disclose to the regulators and the public its activities related to financial and other business performance concerning stakeholders. Each report typically spans hundreds of pages, making it inefficient for human readers to identify and extract the material information.

## Background

Because such reports intend to convey information that is material in impacting stakeholders, such as investors, customers, employees, and the community and environment, human readers often have to identify and extract the relevant information based on its salience to stakeholder relations. As human reading is slow (one to five minutes per page), they often have to use keyword searches to save time. But this approach is inflexible and may miss or misidentify information. For instance, we may define 'community' in the keyword to search for mentions about how the company is contributing to society as part of its social responsibility, but a sentence like "the company received more applicants from community colleges" would be mislabeled as relevant but have nothing to do with "social responsibility". In another example, we may use "customers" or "clients" to search for information related to customer engagement, but a sentence like "the digital company is providing free coding classes to its growing user community" would have been missed in this search.

In the domain of accounting and financial analysis, text mining has been a growing interest since the nineties. The published works have focused on text classification, clustering, information extraction, and text evaluation (for a recent review, see, e.g., Senave, Jans, & Srivastava, 2023). Notably, the existing approaches focus on topic classification/clustering and sentiment analysis, for instance, grouping documents based on their topics or sentiments. Most of such prior works rely on natural language processing (NLP) and machine learning (ML) approaches, which suffer from "being rule-based and inflexible, only considering structured data, or focusing exclusively on numerical cross checking" (Hillebrand et al., 2022: p. 606). The application of pre-trained large language models (LLMs) such as Bidirectional Encoder Representations from Transformers (BERT) and Long Short-Term Memory (LSTM) networks has been fairly limited in the domain (Hillebrand et al., 2022).

Recent years saw some welcome efforts to apply BERT and LSTM models in text mining tasks in accounting and financial analysis. For instance, Hillebrand et al. (2022) introduced a learnable recurrent neural network (RNN)-based pooling mechanism and incorporated domain expert knowledge to label named entities and links related to key performance indicators (KPIs) in business. As another example, Gopalakrishnan, Chen, Dou, Hahn-Powell, Nedunuri, and Zadrozny (2023) fine-tuned SpanBERT and DistilBERT models to identify and extract causes-and-effects relationships from company 10-K reports. Outside the accounting and financial analysis, similar approaches have been applied in biomedical and clinical texts (e.g., Lee et al., 2020), legal documents (e.g., Dyevre, 2021), and material science (e.g., Zhao et al., 2021) and among others.

## Methods

### Data

The data used in this project was based on Gopalakrishnan et al. (2023), in which business graduate students manually annotated and cross-validated 5,962 sentences from 62 financial companies' 2022 annual 10-K reports. The data has a relatively balanced split of positive and negative labels: 3,161 sentences contain material information related to a clear stakeholder group, e.g., investors, customers, employees, local community, and natural environment, whereas 3,401 contain no such information. Here, material information is defined as an explicit statement or discussion about how a company's activity or position may affect the wellbeing or benefits of this stakeholder group.

Below are three example sentences that contain stakeholder-material information, with stakeholder identities added in the brackets:

- "When **a policyholder or insured** [customers] gets sick or hurt, the Company pays cash benefits fairly and promptly for eligible claims." (from Aflac 2022 10-K report).

- "The Corporation is required to maintain a minimum supplementary leverage ratio (SLR) of 3.0 percent plus a leverage buffer of 2.0 percent in order to avoid certain restrictions on **capital distributions and discretionary bonus payments** [investors, shareholders]." (from Bank of America 2022 10-K report).

- "During the 1980s and early 1990s, commercial lines grew as a percentage of our overall business and our exposure to **asbestos and environmental claims** [community, natural environment] grew accordingly." (from Cincinnati Financial Corporation 2022 10-K report).

Below are three example sentences that contain no stakeholder-material information, with a note explaining why it was labeled as non-material information to any stakeholder influence:

- "$114 million of coverage in effect for 2021 does not include an exclusion for communicable diseases such as a virus." (from Cincinnati Financial Corporation 2022 10-K report). Note: this sentence talks about the policy coverage criteria, and has no clear indication of how it may affect any stakeholders.

- "When public quotations are not available, because of the highly liquid nature of these assets, carrying amounts may be used to approximate fair values, which are reflected in Level 2." (from Principal Financial Group Inc 2022 10-K report). Note: this sentence talks about how asset values are calculated and has no clear indication of how it may affect any stakeholders.

- "These forward-looking statements involve certain risks and uncertainties which could cause actual results to differ materially." (from State Street Corp 2022 10-K report). Note: this sentence seems to talk about firm outcome results, but has no indicator to which stakeholder the results may affect.

**Baseline model**

The baseline model is to simulate how human readers may use keyword searches to identify the sentences that may likely carry stakeholder-material information. A comprehensive list of keywords that may indicate stakeholder mentions include:

- Customers: "customer*", "client*", "consumer*";
- Investors: "investor*", "financ*", "shareholder*", "stockholder*", "owners*", "investment*", "return on*", "net income*", "profit*", "revenue*", "earnings*";
- Employees: "employee*", "worker*", "manager*";
- Community/Natrual environment: "society", "societal", "social responsib*", "social performance", "communit*", "natrual environment*", and "ecolog*"

* represents any trailing character(s) in order to take into account any pluralistic or adjective terms.

As keywords need to be pre-defined, this approach is rule-based and inflexible. It is hard to pre-populate an exhaustive list as different companies may use different expressions to describe their stakeholders, and some stakeholder-alike expressions may have nothing to do with a company's stakeholders. For instance, a sentence like "More community colleges are preparing their graduates for a job in the financial industry" is not talking about a firm's impact on the community but a general description of the job market make-up. As another example, some firms may use uncommon terms to describe their consumers such as "policyholders", "insured", or "beneficiaries", which may not be pre-captured by a generic list of keywords (as in one of the aforementioned examples). A lot of domain expert judgments need to be made when reading a report to identify whether a sentence is describing material information related to a stakeholder.

**Fine-tuned BERT and LSTM models**

I now fine-tuned BERT models and RNN models with LSTM layers in the domain of text classification and then compared their performance against the baseline model and each other in the test data.

BERT is a pre-trained deep learning model that uses transformer architecture to generate contextually rich word embeddings, revolutionizing natural language understanding tasks. Specifically, I have experimented with five of the most used BERT models, including BERT-en-uncased (Devin et al., 2018), ALBERT-en-base (Lan et al., 2019), BERT-experts-wiki-books (Google., 2023), BERT-talking-heads-base (Shazeer et al., 2020), and DistilBERT-en-uncased (Sanh et al., 2019).

An RNN language model with LSTM is a type of deep learning model that processes sequential data, like sentences, by maintaining a memory mechanism that effectively captures long-range dependencies in the input, enabling it to better understand and generate coherent natural language text. LSTM is specifically designed to address the vanishing gradient problem in traditional RNNs, making it better at handling long sequences and preserving essential information over time. Specifically, I have experimented with two RNN models, including an RNN with one LSTM layer, followed by an RNN with a stack of two LSTM layers (Liu et al., 2016).

Specifically, I fine-tune each pre-trained model to do text classification and compare the performance metrics against the baseline model which is based on keyword search. Figure 1 shows the workflow of the task. First, the labeled data were downloaded and preprocessed. Second, the labeled data were randomly split into 80% train data and 20% test/validation data. Third, a pre-trained BERT or RNN model was compiled on the train data, and all weights were retrained in each epoch. Due to the GPU usage limit, BERT model fine-tuning was limited by five epochs. Whereas run on CPUs, the RNN models were fine-tuned in 10 epochs. Finally, the best model among all epochs was selected in the end based on the greatest F1 score on the test/validation data.

**Figure 1. Model Pipeline**

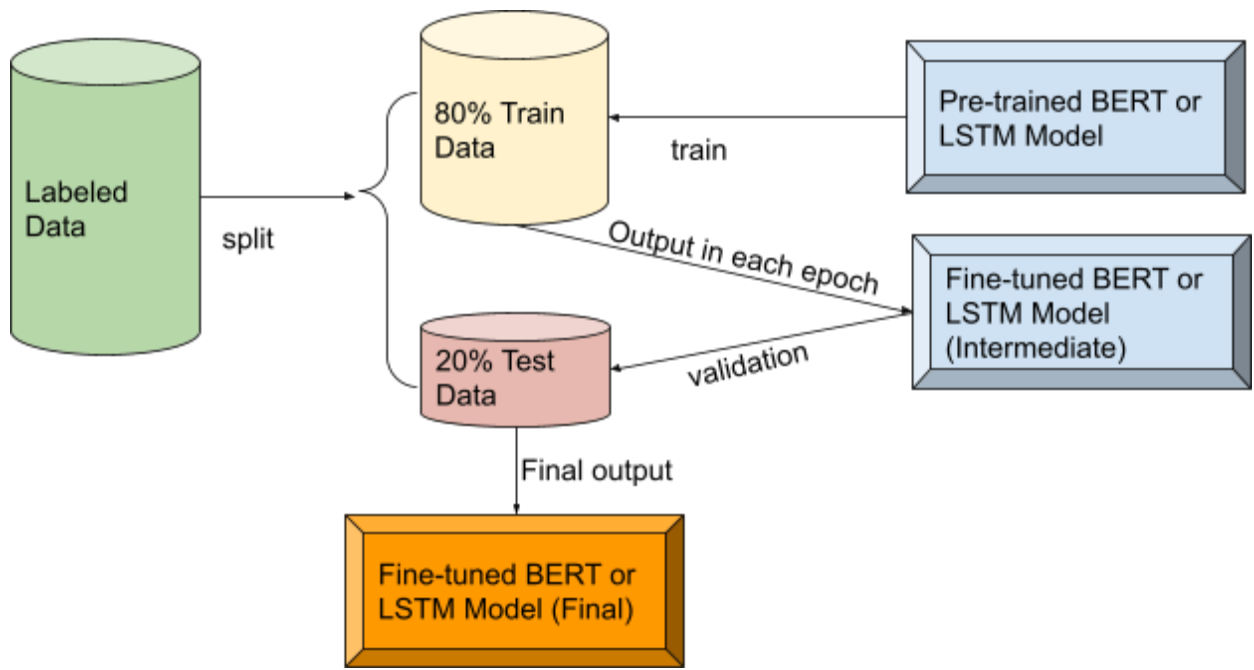

Note: See Appendix Figure 1 for a detailed compilation of BERT fine-tuning, and Appendix Figures 1B for a detailed compilation of RNN LSTM fine-tuning.

## Results and Discussion

*Stakeholder-Material Information without Distinguishing Stakeholder Identities*

Model performance metrics were collected and compared in Table 1, including accuracy, recall, precision, and F1 score on the test data. Specifically, because the task aims to identify stakeholder-material texts and exclude non-material texts from a document, one should strike a good balance among all the performance metrics. Because the training data is highly imbalanced, with the majority being negative labels (i.e., nonmaterial information), the F1 score was used as the primary metric, representing how well the fine-tuned model predicted the right labels. But all other metrics were also considered secondary metrics.

Table 1. Comparison in Performance Metrics

| Model | Accuracy | Recall | Precision | F1 |
|---|---|---|---|---|
| rule-based (baseline) | 0.781 | 0.677 | 0.837 | 0.749 |
| BERT-en-uncased | 0.894 | **0.898** | 0.888 | 0.893 |
| ALBERT-en-base | **0.904** | 0.873 | **0.927** | **0.899** |
| BERT-experts-wiki-books | 0.864 | 0.818 | 0.896 | 0.856 |
| BERT-talking-heads-base | 0.900 | 0.887 | 0.908 | 0.897 |
| distilBERT-en-uncased | 0.896 | 0.887 | 0.901 | 0.894 |
| RNN (1 LSTM layer) | 0.816 | 0.755 | 0.854 | 0.801 |
| RNN (2 LSTM layers) | 0.807 | 0.727 | 0.859 | 0.787 |

Notes: Please see the Appendix for the performance metrics of train vs. test/validation data over epochs.

First, as hypothesized earlier, the rule-based baseline model using keyword searches did not perform well due to its inflexibility and incomplete inclusion. Specifically, it has a relatively poor F1-score of 0.749, an accuracy of 0.781, a recall of 0.677, and a precision of 0.837, suggesting that it has made a lot of errors in terms of both false positives and false negatives.

Overall, ad Table 1 reports, fine-tuned model using ALBERT-en-base stood out as the best-performing model, with the highest F1 score of 0.899, the highest accuracy of 0.904, the highest precision of 0.927, and close to the best recall of 0.873. In other words, one may gain an additional 0.123 accuracy and 0.15 F1 score from the best fine-tuned BERT model vis-a-vis the keyword searches. The standard BERT-en-uncased model also performs very well, with the highest recall of 0.898, and close to best accuracy, precision, and F1.

Some less-performing models are also worth mentioning. First, initially, I expected that the BERT-experts-wiki-books would be a strong contender to be the best performer, as it drew on expert data from Wikipedia, rather than any website sources. But it turns out the Wikipedia data may not well represent specialized knowledge domains, such as financial reports in our case. Second, overall, pre-trained BERT models do perform better than the LSTM models, suggesting that our task may have benefited from pre-training from a massive corpus of text data and context awareness. Also, BERT was trained to understand the context of words than LSTM.

*Stakeholder-Material Information Distinguishing Stakeholder Identities*

Next, I replicated the model development above for each of the four stakeholder identities, including customers (CUS), investors (INV), employees (EMP), and the community and natural environment (SOC). Table 2 below shows the baseline performance metrics by each stakeholder identity using the rule-based keyword searches. Unlike the work above, for each of the four stakeholder identities, the training sample is highly imbalanced, where the majority of sentences were labelled negative. For instance, out of 5,962 labelled sentences in the data, investor-material sentences represented only 33.95%, followed by customer-material ones with 10.37%, employee-material ones with 3.30%, and community and natural environment-meterial ones with only 0.35%. Because of this high imbalance, F1-score was the focus to determine the performance of a fine-tuned or retrained model.

**Table 2. Performance Metrics of Rule-based Approach**

| Stakeholder | Accuracy | Recall | Precision | F1 | Number of positive labels | % among all labelled sentences |
|---|---|---|---|---|---|---|
| CUS | 0.982 | 0.884 | 0.935 | 0.909 | 618 | 10.366 |
| INV | 0.838 | 0.712 | 0.797 | 0.752 | 2024 | 33.948 |
| EMP | 0.982 | 0.627 | 0.964 | 0.760 | 197 | 3.304 |
| SOC | 0.993 | 0.297 | 0.905 | 0.447 | 21 | 0.352 |

Note: The numbers of positive labels do not add up to the total number of stakeholder-material sentences for two reasons. First, the same sentence may contain material information for multiple stakeholder identities. Second, a sentence may contain material information for a stakeholder, but it is not clear to which specific stakeholder the information is relevant to. Such a sentence was labelled a positive earlier as being stakeholder-material, but labelled as negative for any specific stakeholder identity.

Tables 3A to 3D reports the performance metrics for fine-tuned BERT models and LSTM models vis-a-vis the baseline. Similar to the findings in the overall performance without distinguishing different stakeholder groups, fine-tuned BERT models perform significantly better than the baseline, and outperformed the LSTM models in all cases.

Specifically, for predicting customer-material sentences, fine-tuned models based on BERT-en-uncased, BERT-talking-heads-base, and distilBERT-en-uncased reached the highest F1-score (0.97, compared to the 0.909 baseline). For predicting investor-material sentences, fine-tuned distilBERT-en-uncased models reported the greatest F1-score (0.915, significantly higher than the 0.752 baseline). For predicting employee-material sentences, fine-tuned BERT-talking-heads-base model reported the highest F1-score (0.847, compared to a baseline of 0.76). Finally, for predicting community/natural environment-material sentences, fine-tuned BERT-en-uncased model reported the highest F1-score of 0.588. While this score is still low, it is a significant improvement from the 0.447 baseline. A low F1-score is a result of the limited training data in the sample.

**Table 3A. Comparison in Performance Metrics on Customer Information**

| Model | Accuracy | Recall | Precision | F1 |
|---|---|---|---|---|
| rule-based (baseline) | 0.982 | 0.884 | 0.935 | 0.909 |
| BERT-en-uncased | **0.994** | 0.970 | **0.970** | **0.970** |
| ALBERT-en-base | 0.991 | **0.977** | 0.935 | 0.956 |
| BERT-experts-wiki-books | 0.992 | **0.977** | 0.949 | 0.963 |
| BERT-talking-heads-base | **0.994** | 0.970 | **0.970** | **0.970** |
| distilBERT-en-uncased | **0.994** | 0.970 | **0.970** | **0.970** |
| RNN (1 LSTM layer) | 0.954 | 0.579 | 0.951 | 0.720 |
| RNN (2 LSTM layers) | 0.964 | 0.722 | 0.906 | 0.803 |

**Table 3B. Comparison in Performance Metrics on Investor Information**

| Model | Accuracy | Recall | Precision | F1 |
|---|---:|---:|---:|---:|
| rule-based (baseline) | 0.838 | 0.712 | 0.797 | 0.752 |
| BERT-en-uncased | 0.938 | 0.915 | 0.911 | 0.913 |
| ALBERT-en-base | 0.937 | **0.921** | 0.904 | 0.913 |
| BERT-experts-wiki-books | 0.935 | 0.894 | 0.923 | 0.908 |
| BERT-talking-heads-base | 0.937 | 0.900 | 0.922 | 0.911 |
| **distilBERT-en-uncased** | **0.940** | 0.904 | **0.926** | **0.915** |
| RNN (1 LSTM layer) | 0.866 | 0.783 | 0.833 | 0.807 |
| RNN (2 LSTM layers) | 0.872 | 0.781 | 0.850 | 0.814 |

**Table 3C. Comparison in Performance Metrics on Employee Information**

| Model | Accuracy | Recall | Precision | F1 |
|---|---:|---:|---:|---:|
| rule-based (baseline) | 0.982 | 0.627 | **0.964** | 0.760 |
| BERT-en-uncased | 0.977 | 0.809 | 0.764 | 0.786 |
| ALBERT-en-base | 0.979 | 0.706 | 0.857 | 0.774 |
| BERT-experts-wiki-books | 0.978 | 0.838 | 0.760 | 0.797 |
| **BERT-talking-heads-base** | **0.984** | **0.853** | 0.841 | **0.847** |
| distilBERT-en-uncased | 0.976 | 0.824 | 0.737 | 0.778 |
| RNN (1 LSTM layer) | 0.963 | 0.485 | 0.702 | 0.574 |
| RNN (2 LSTM layers) | 0.960 | 0.397 | 0.692 | 0.505 |

**Table 3D. Comparison in Performance Metrics on Community/Environment Information**

| Model | Accuracy | Recall | Precision | F1 |
|---|---:|---:|---:|---:|
| rule-based (baseline) | 0.993 | 0.297 | 0.905 | 0.447 |
| **BERT-en-uncased** | **0.995** | **0.417** | **1.000** | **0.588** |
| ALBERT-en-base | 0.993 | 0.250 | **1.000** | 0.400 |
| BERT-experts-wiki-books | 0.993 | **0.417** | 0.714 | 0.526 |
| BERT-talking-heads-base | 0.994 | **0.417** | 0.833 | 0.556 |
| distilBERT-en-uncased | 0.992 | 0.250 | 0.750 | 0.375 |
| RNN (1 LSTM layer) | 0.991 | 0.000 | 0.000 | 0.000 |
| RNN (2 LSTM layers) | 0.991 | 0.000 | 0.000 | 0.000 |

## Conclusion

In this project, I seek to solve the reading inefficiency problem concerning professionals such as financial analysts, auditors, regulators, accountants, and managers who need to efficiently capture the material information from company annual reports. The current method based on key searches turns out to be very inaccurate, generating both very high false positives and false negatives. In this project, I drew on recently published training data on financial companies' 10-K reports and experimented with five popular BERT models and two LSTM models. The final best model has increased the F1 score of finding material information from texts by 15%. The checkpoint of the best model can be potentially productized as a tool for future analysts to quickly identify the key information from company reports.

Meanwhile, there are several ways to extend this research. First, the current project is built on a very specific industry, i.e., financial companies, and a generalization of the model would require the collection of more data across other industries. Second, the current data is labeled manually by business domain experts, which was costly. We would benefit from more training data, e.g., more recent years of publications, and more fine-grained classification of stakeholders and types of impact. It is worth experimenting with a human-GPT collaboration to create a large training dataset for this job. Finally, while the performance of fine-tuned models is significantly better than the baseline using keyword searches, it requires further improvement to catch up with human readers.

Appendix Figure 1A. Classifier Model using BERT

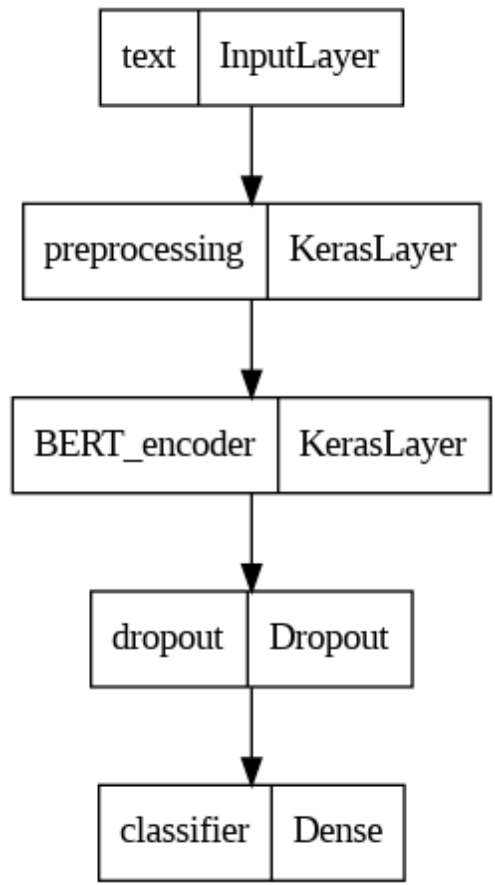

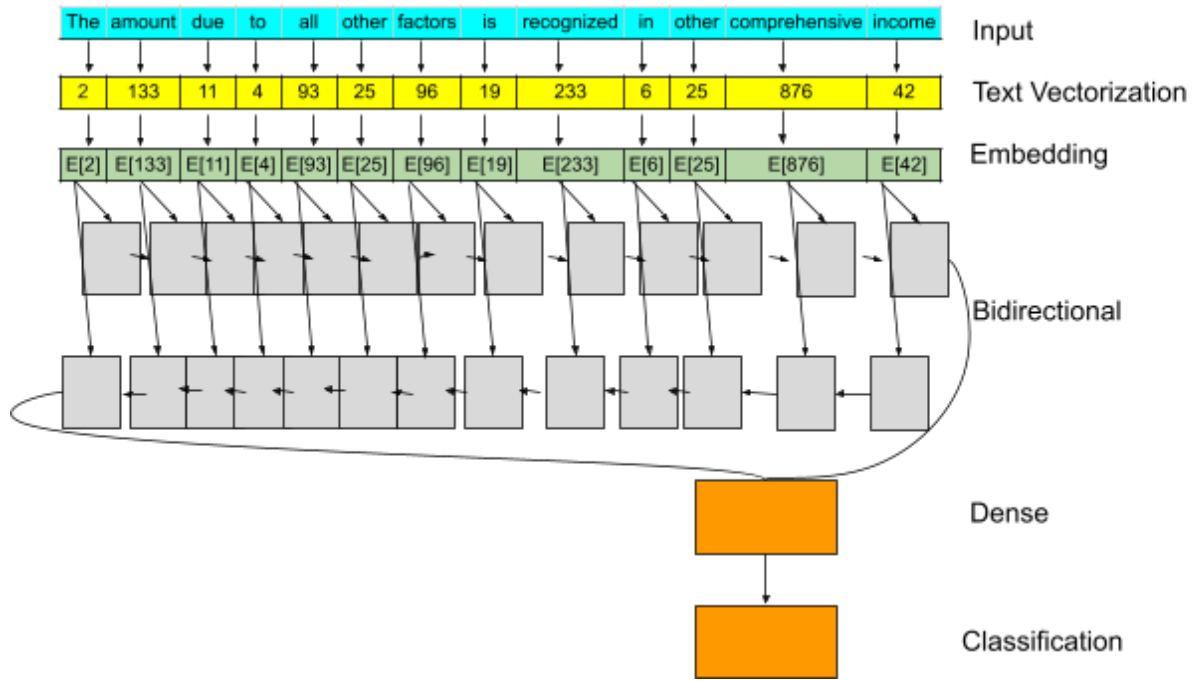

Appendix Figure 1B. Classifier Model using RNN

Appendix Figure 2A. Performance metrics over epochs in fine-tuning bert-en-uncased

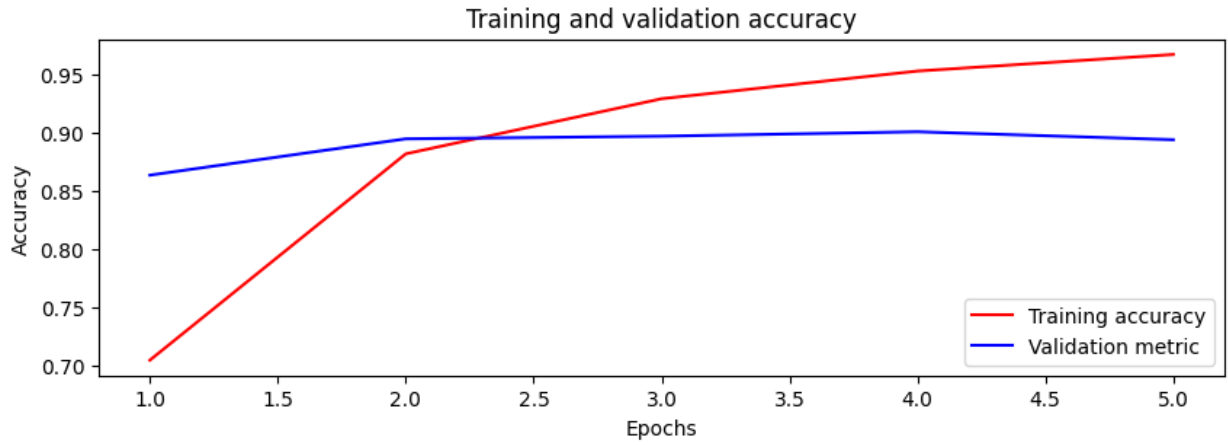

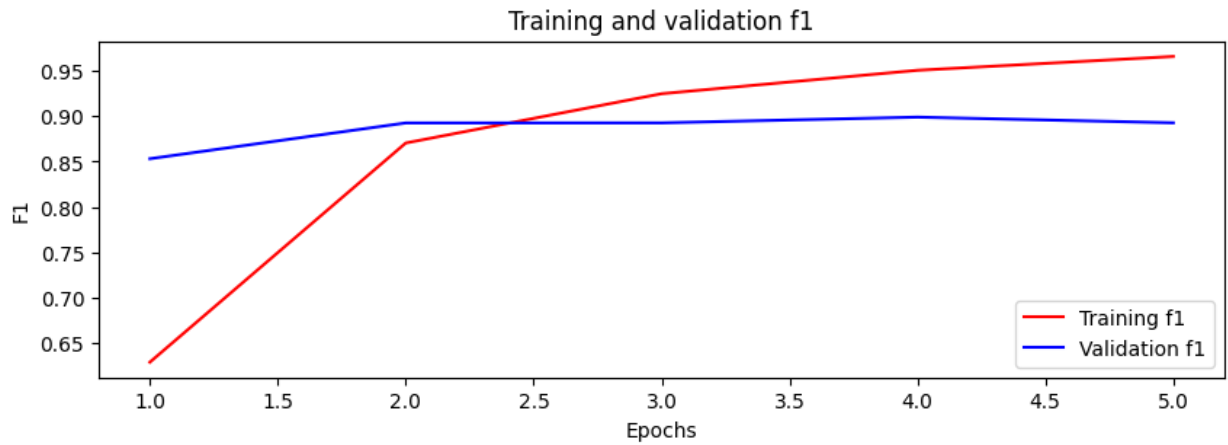

Appendix Figure 2B. Performance metrics over epochs in fine-tuning albert-en-base

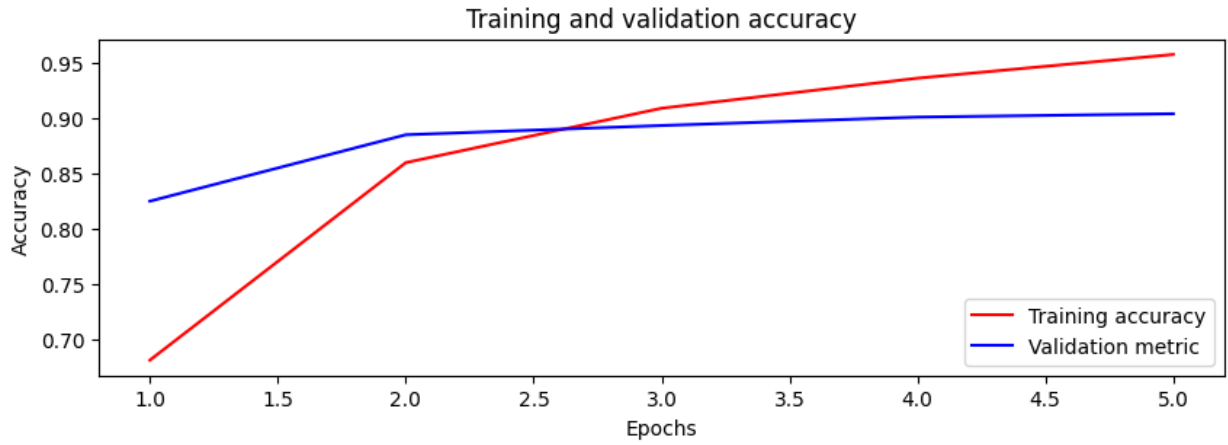

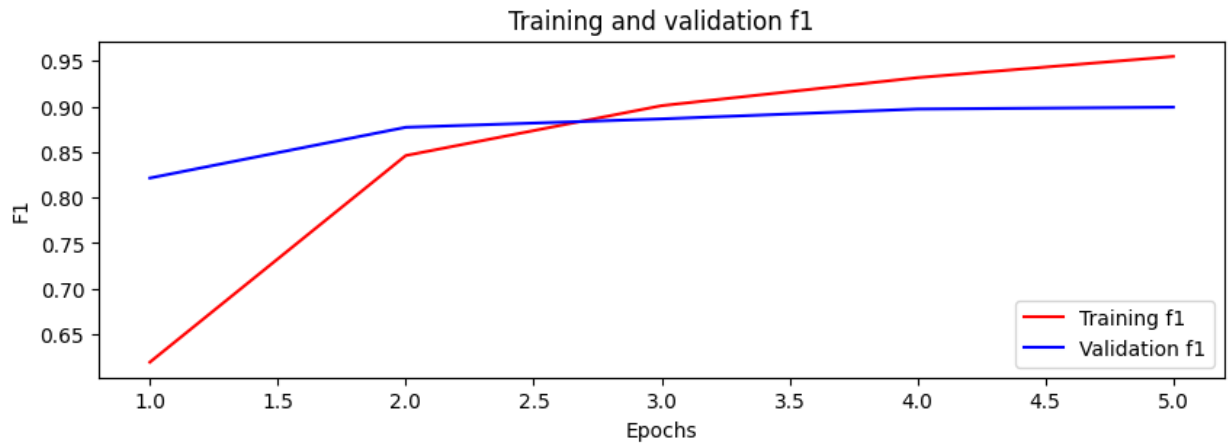

Appendix Figure 2C. Performance metrics over epochs in fine-tuning experts-wiki-books

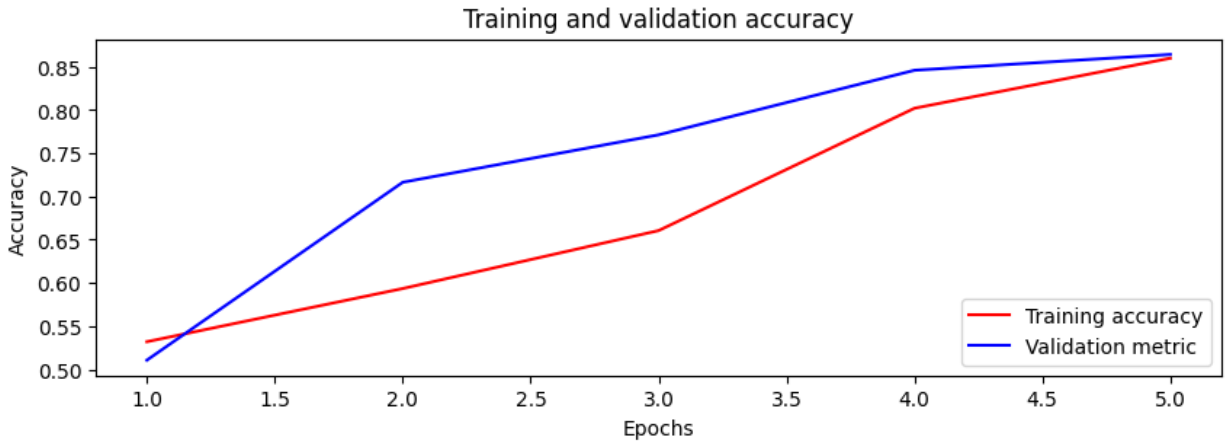

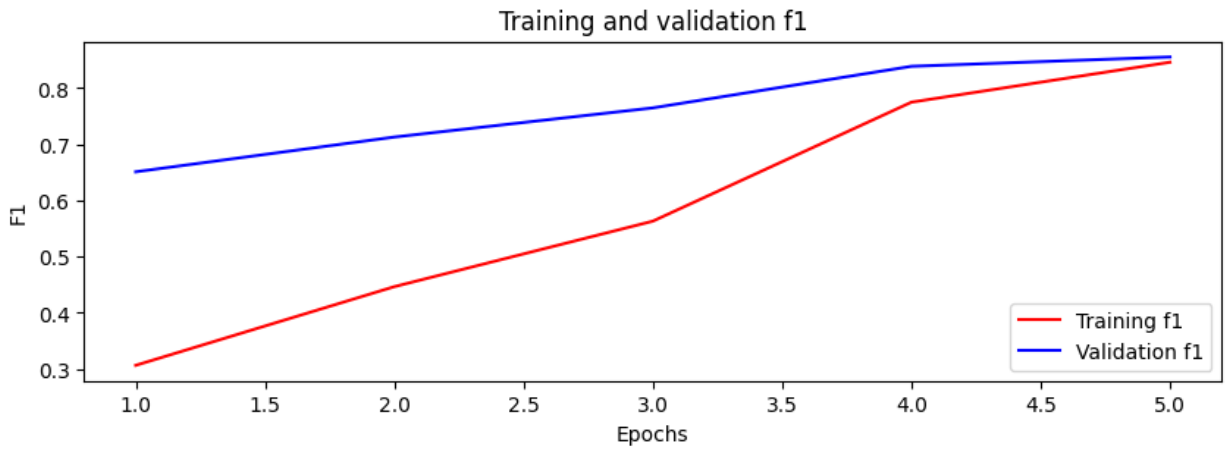

Appendix Figure 2D. Performance metrics over epochs in fine-tuning talking-heads-base

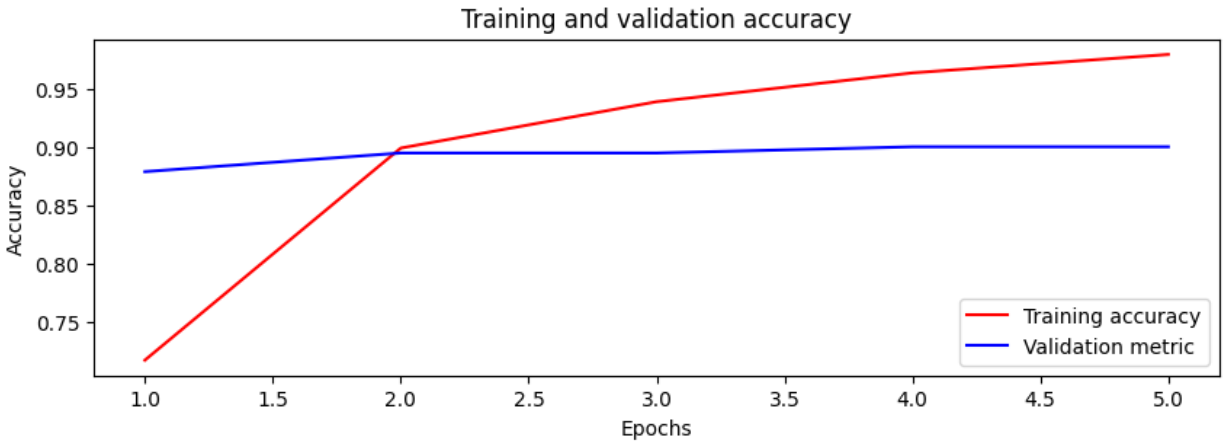

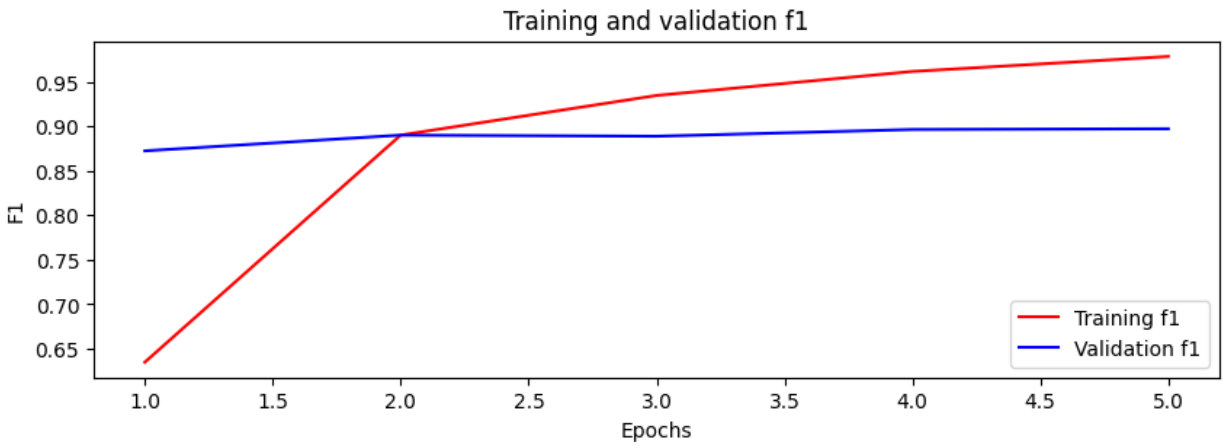

Appendix Figure 2E. Performance metrics over epochs in fine-tuning distilbert-en-uncased

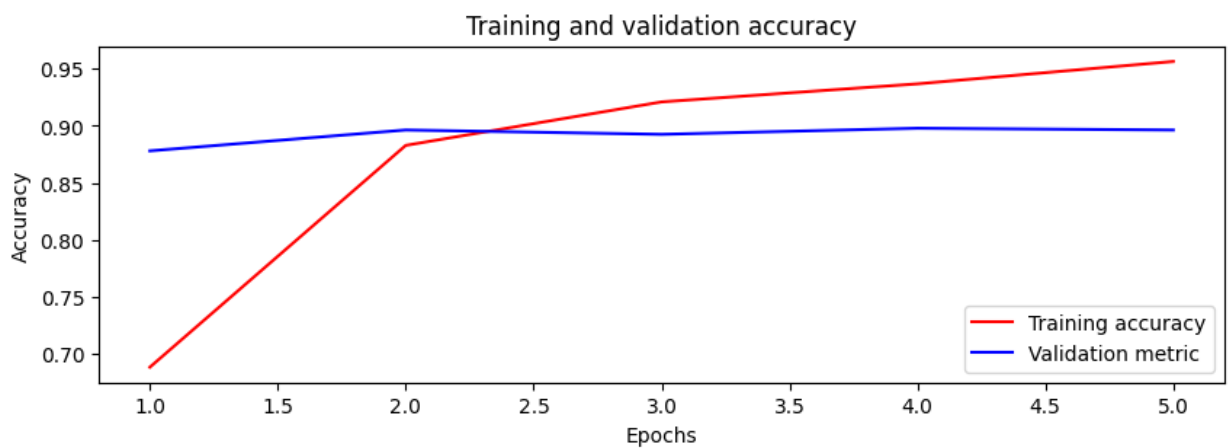

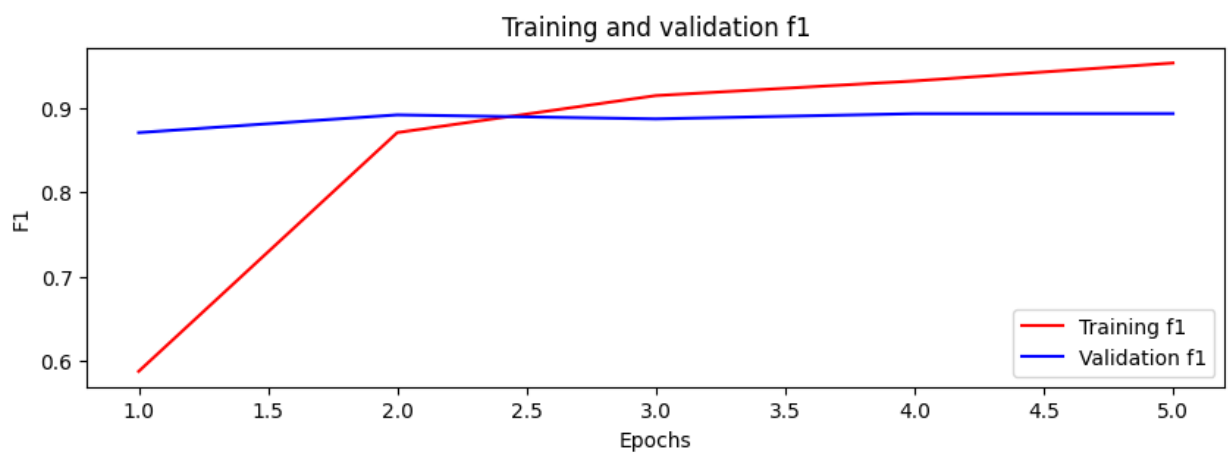

Appendix Figure 2F. Performance metrics over epochs in training RNN (1 LSTM layer)

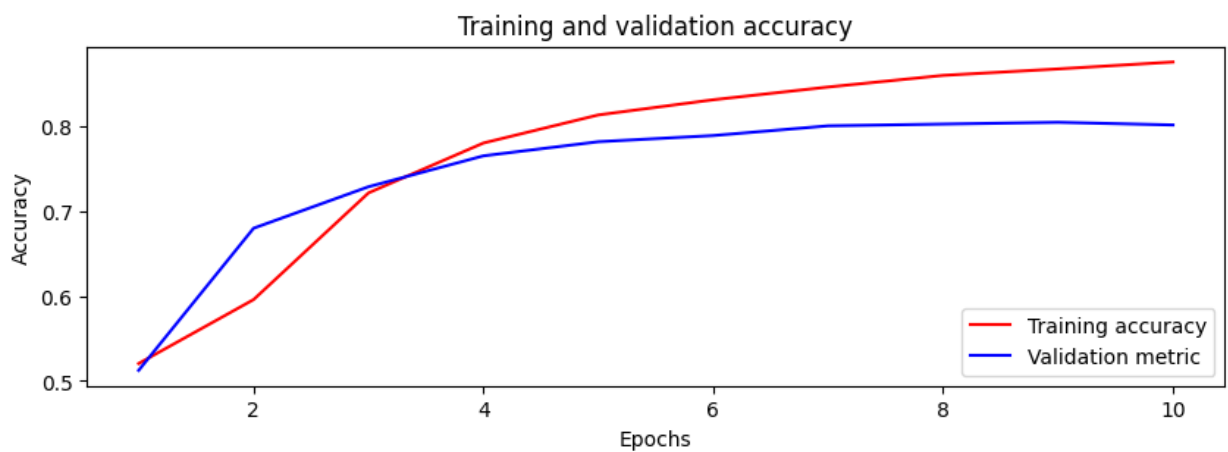

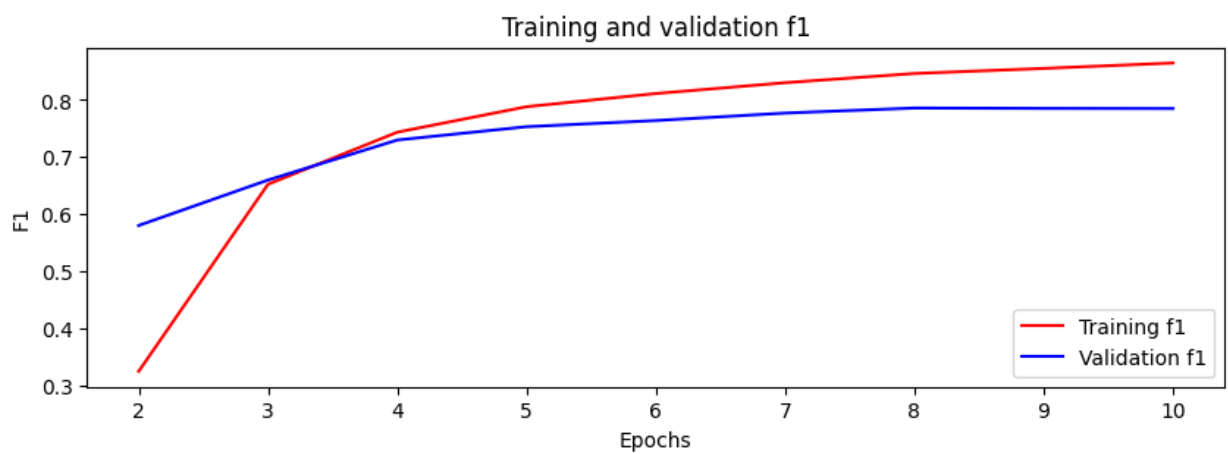

Appendix Figure 2G. Performance metrics over epochs in training RNN (2 LSTM layers)

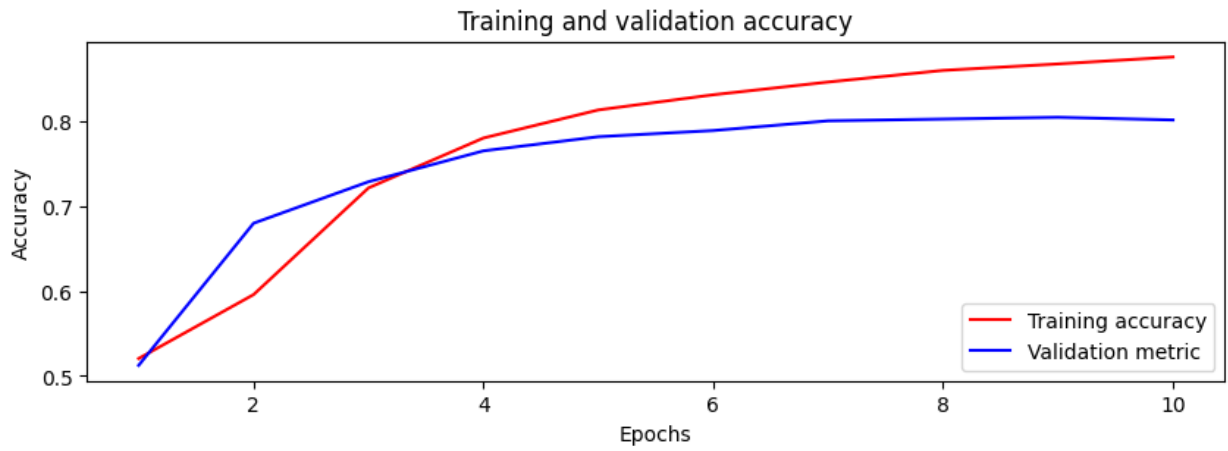

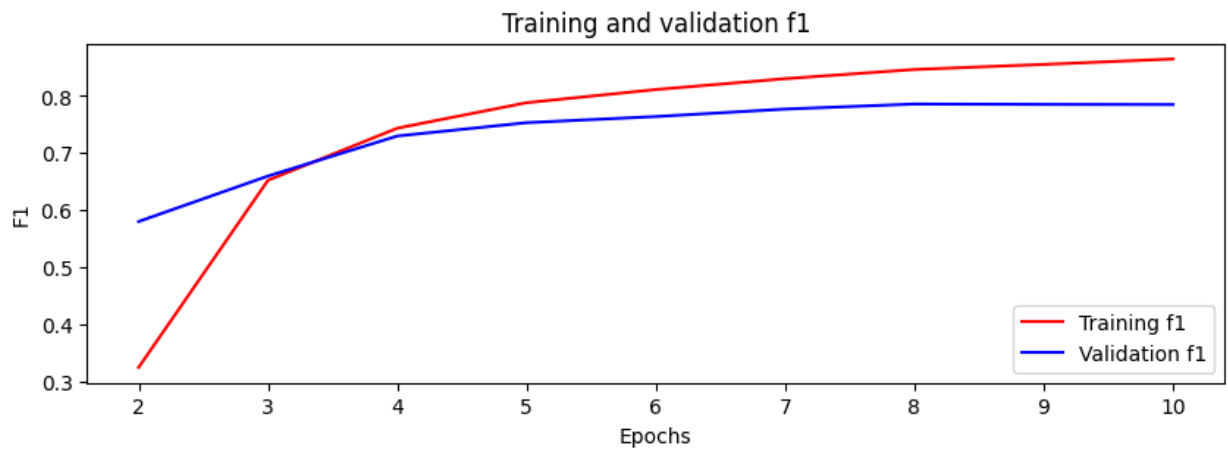